\documentclass[10pt,twocolumn,letterpaper]{article}

\usepackage{cvpr}      

\usepackage{graphicx}
\usepackage{amsmath}
\usepackage{amssymb}
\usepackage{booktabs}
\usepackage{multicol}
\usepackage{multirow}
\usepackage{booktabs}
\usepackage{tabularx}
\usepackage{url}            
\usepackage{booktabs}       
\usepackage{amsfonts}       
\usepackage{nicefrac}       
\usepackage{microtype}      
\usepackage{xcolor}         
\usepackage{graphicx}
\usepackage{amsmath}
\usepackage{bm}
\usepackage{multirow}
\usepackage{tabu}
\usepackage{siunitx}
\usepackage{adjustbox}
\usepackage{subcaption}
\usepackage{siunitx}
\usepackage{makecell}
\usepackage{xcolor}
\usepackage{colortbl}
\usepackage{color}
\usepackage{subcaption}
\usepackage{pifont}
\newcommand{\cmark}{\ding{51}}%
\newcommand{\xmark}{\ding{55}}%
\usepackage{multirow}
\definecolor{Gray}{gray}{0.95}
\definecolor{Cyan}{rgb}{0.88,1,1}
\usepackage[pagebackref,breaklinks,colorlinks]{hyperref}
\newcommand{\paragrapha}[2][5pt]{\vspace{#1}\noindent\textbf{#2}}

\usepackage[capitalize]{cleveref}
\crefname{section}{Sec.}{Secs.}
\Crefname{section}{Section}{Sections}
\Crefname{table}{Table}{Tables}
\crefname{table}{Tab.}{Tabs.}

\DeclareMathOperator{\mhsa}{MHSA}
\DeclareMathOperator{\transdecoder}{TransDecoder}
\DeclareMathOperator{\ce}{CrossEntropy}
\DeclareMathOperator{\bce}{BinaryCrossEntropy}
\DeclareMathOperator{\sigmoid}{Sigmoid}
\DeclareMathOperator{\softmax}{Softmax}

\def\cvprPaperID{1965} 
\def\confName{CVPR}
\def\confYear{2022}

\definecolor{Gray}{gray}{0.95}
\newcommand\cb[1]{\text{\color{blue} #1}}

\begin{document}

\title{DenseCLIP: Language-Guided Dense Prediction with Context-Aware Prompting}

\author{
  Yongming Rao\thanks{Equal contribution. ~~\textsuperscript{\dag}Corresponding author.} $  ^{,1}$,
  Wenliang Zhao$^{*,1}$, 
  Guangyi Chen$^{1}$,
  Yansong Tang$^{1}$, \\
  Zheng Zhu$^{1}$,
  Guan Huang$^{2}$,
  Jie Zhou$^{1}$, 
  Jiwen Lu$^{\dagger,1}$      \\
    $^1$Tsinghua University ~~ $^2$PhiGent Robotics\\
  }
  
\maketitle

\begin{abstract}
Recent progress has shown that large-scale pre-training using contrastive image-text pairs can be a promising alternative for high-quality visual representation learning from natural language supervision. Benefiting from a broader source of supervision, this new paradigm exhibits impressive transferability to downstream classification tasks and datasets. However, the problem of transferring the knowledge learned from image-text pairs to more complex dense prediction tasks has barely been visited. In this work, we present a new framework for dense prediction by implicitly and explicitly leveraging the pre-trained knowledge from CLIP. Specifically, we convert the original image-text matching problem in CLIP to a pixel-text matching problem and use the pixel-text score maps to guide the learning of dense prediction models. By further using the contextual information from the image to prompt the language model, we are able to facilitate our model to better exploit the pre-trained knowledge. Our method is model-agnostic, which can be applied to arbitrary dense prediction systems and various pre-trained visual backbones including both CLIP models and ImageNet pre-trained models. Extensive experiments demonstrate the superior performance of our methods on semantic segmentation, object detection, and instance segmentation tasks. Code is available at \url{https://github.com/raoyongming/DenseCLIP}.
\end{abstract}

\section{Introduction}

\begin{figure}[t]
  \centering
  \includegraphics[width=0.99\linewidth]{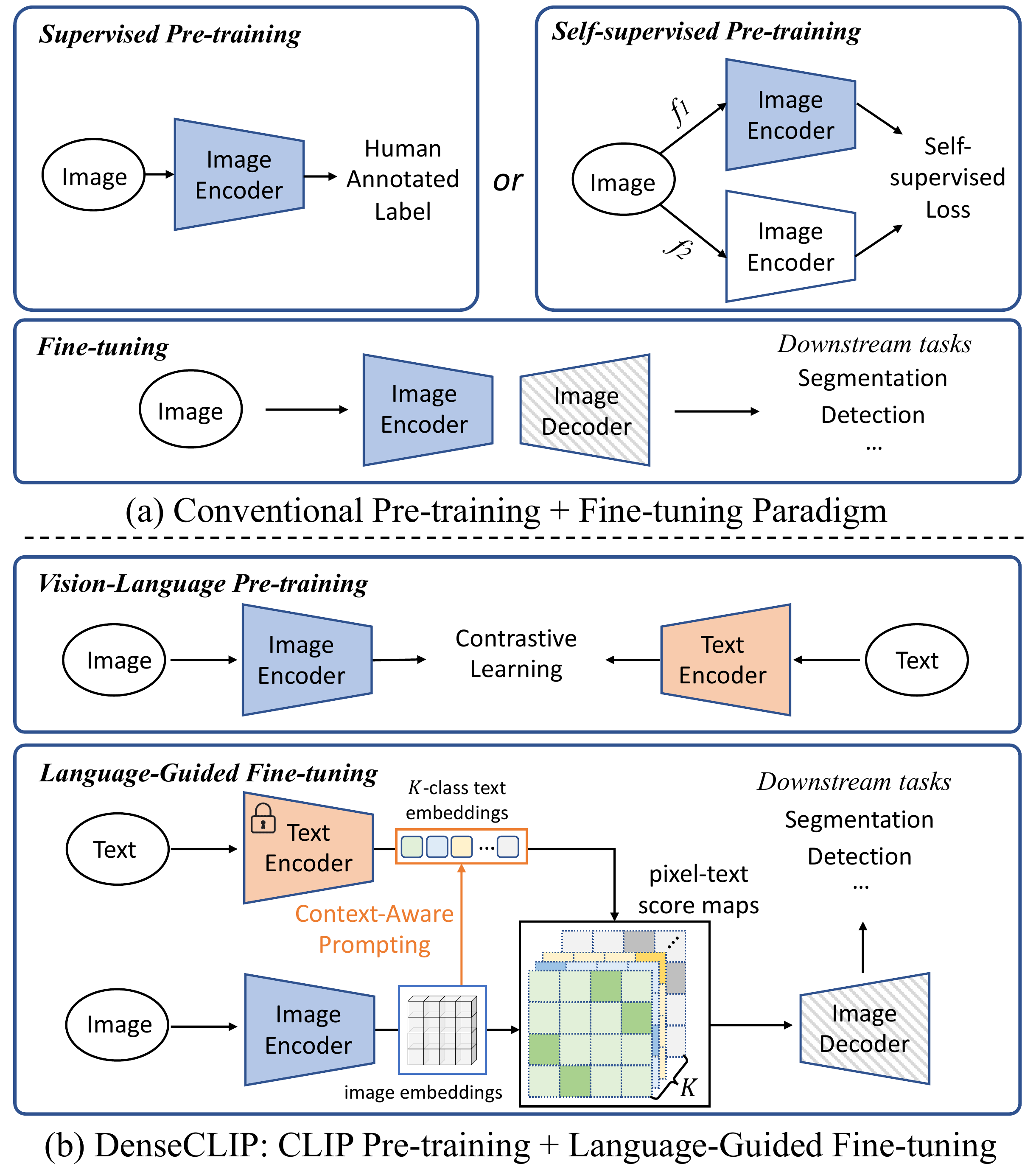}
   \caption{ Comparisons of the conventional ``pre-training + fine-tuning'' paradigm and our proposed \emph{DenseCLIP}. The pre-training + fine-tuning paradigm directly applies the image pre-trained model as the initialization of encoder. Differently, DenseCLIP transfers the knowledge learned with image-text contrastive learning to dense prediction models by introducing a new pixel-text matching task and further using the contextual information from images to prompt pre-trained language model.}
   \label{fig:top}
   \vspace{-10pt}
\end{figure}

The ``\emph{pre-training + fine-tuning}" paradigm is recognized as one of the key discoveries that has largely pushed the state-of-the-art for various downstream computer vision tasks, including  image classification~\cite{krizhevsky2012imagenet,bit,vit}, object detection~\cite{girshick2014rich,ren2015faster}, semantic segmentation~\cite{fcn,chen2017deeplab}, and action recognition~\cite{carreira2017quo}. Due to the high annotation and computation cost of the per-pixel prediction, pre-training is even more critical for dense prediction tasks. As illustrated in Figure~\ref{fig:top} (a), the pre-training step is usually accomplished via supervised classification or self-supervised learning of the backbone model on large-scale datasets like ImageNet~\cite{deng2009imagenet}. Then, a task-specific module like a detector or a segmentation decoder is added to the backbone and the whole model is fine-tuned on the target dataset with less training data~\cite{chen2017deeplab,ren2015faster}. 


Different from conventional supervised and self-supervised pre-training methods only based on images, Contrastive Language-Image Pre-training (CLIP)~\cite{clip} is a new framework to learn high-quality visual representation by exploring contrastive learning with large-scale noisy image-text pairs. By exploiting the semantic relationships between the images and the associated texts, this new framework benefits from rich and semantic level supervision from texts while enjoying a broader and cheaper source of data. Thanks to the language supervision, models pre-trained via CLIP achieve impressive results on various visual classification tasks with no or very limited annotations~\cite{wang2021actionclip,coop,clip-adapter}.

\begin{figure}[t]
  \centering
  \includegraphics[width=\linewidth]{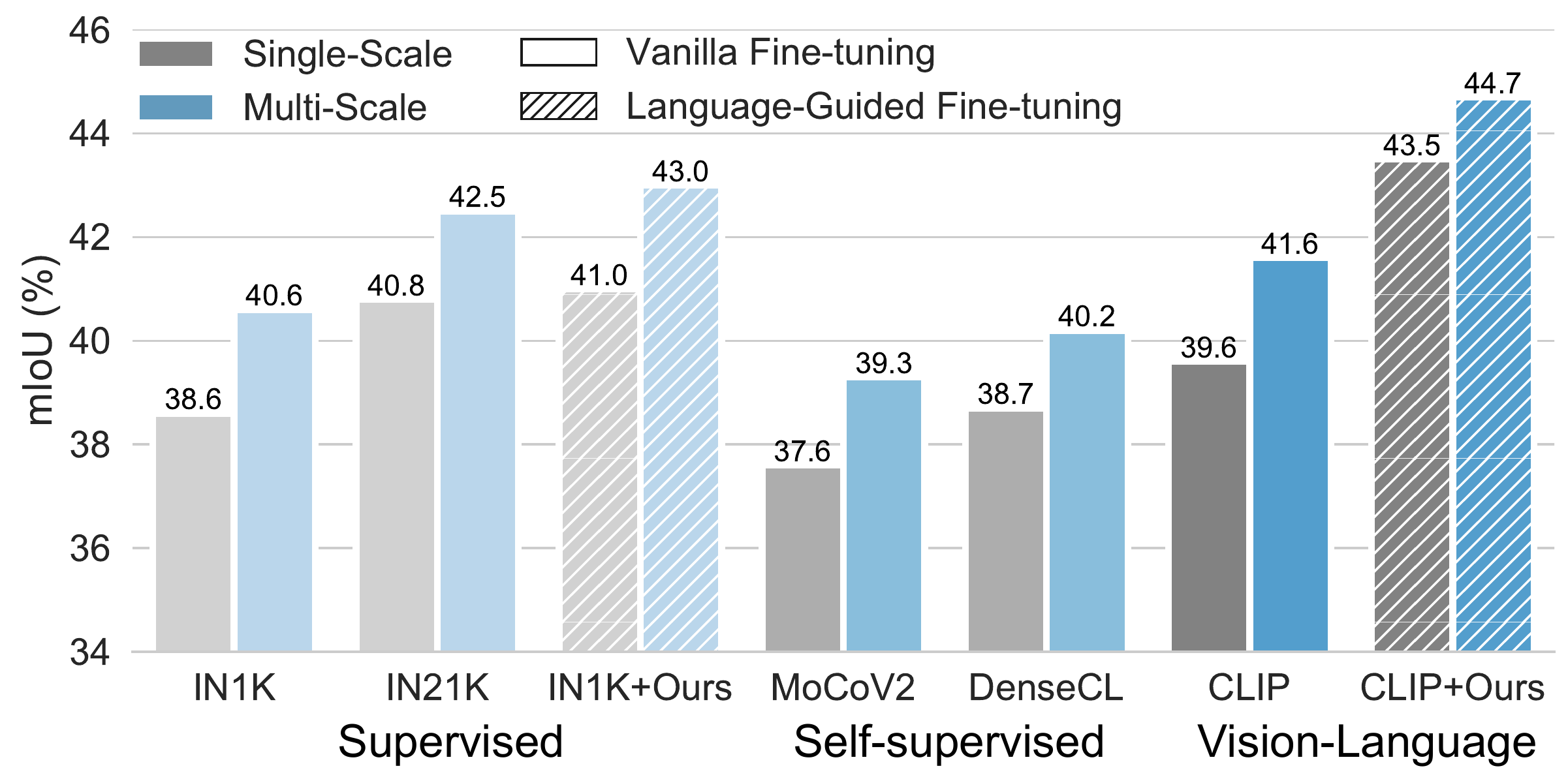}
   \caption{Results of different pre-training and fine-tuning strategies on the semantic segmentation task. We report the single-scale and multi-scale mIoU on ADE20K~\cite{ade} of different pre-trained ResNet-50~\cite{he2016deep} models, including supervised ImageNet1K~\cite{deng2009imagenet} (IN1K) and ImageNet21K~\cite{deng2009imagenet,ridnik2021imagenet} (IN21K), self-supervised MoCoV2~\cite{mocov2} and DenseCL~\cite{wang2021dense}, and vision-language model CLIP. Equipped with DenseCLIP, we show that large-scale vision-language pre-training can substantially improve the dense prediction performance (+4.9\%/+4.1\%) over the commonly used ImageNet pre-training. }    \vspace{-10pt}
   \label{fig:pre-trains}
\end{figure}

Very recently, several efforts have been made to adopt the prompt engineering from NLP community~\cite{liu2021pre} to better transfer the CLIP models to the downstream visual classification tasks. Several learning-based prompting methods~\cite{coop,yao2021cpt,clip-adapter,zhang2021tip} are proposed to modify the output of the language model to better adapt to the new tasks. However, they mainly focus on transferring the CLIP model to classification tasks by performing image-text matching, which is much close to the original pre-training task. The problem of transferring the knowledge learning from image-text pairs to more complex dense prediction tasks and a more generic setting has barely been visited.

In this paper, we study how to fine-tune the pre-trained CLIP models to dense prediction tasks. Compared to conventional ImageNet pre-trained models, 
one distinct challenge is the gap between the upstream contrastive pre-training task and the downstream per-pixel prediction task, where the former involves instance-level representation of both images and texts, and the latter is only based on the visual information at the pixel level.
To tackle this problem, we present a new language-guided dense prediction framework named \emph{DenseCLIP}. As shown in Figure~\ref{fig:top} (b), it is designed for various \emph{Dense} prediction tasks by implicitly and explicitly leveraging the pre-trained knowledge from \emph{CLIP} models.
An implicit way to exploit the pre-trained knowledge is to directly fine-tune the models on  the downstream datasets. Our results show that the CLIP models can outperform the conventional ImageNet pre-trained models with some modifications on hyper-parameters (see the CLIP result in Figure~\ref{fig:pre-trains}). 
But the straightforward way cannot fully exploit the potential of the CLIP models. Inspired by the original contrastive learning framework in CLIP, 
we propose to convert the original image-text matching problem in CLIP to a pixel-text matching problem and use the pixel-text score maps to guide the learning of dense prediction models explicitly. By further using the contextual information from the image to prompt the language model with a Transformer~\cite{transformer} module, we are able to facilitate our model to better exploit the pre-trained knowledge by optimizing the text embeddings. 

Our method can be a plug-and-play module to improve the fine-tuning of CLIP pre-trained models on off-the-shelf dense prediction methods and tasks. By applying our method to the popular semantic segmentation framework semantic FPN~\cite{semfpn} on the challenging ADE20K~\cite{ade} dataset, we exhibit +4.9\%, +4.7\% and +2.3\% mIoU improvement compared over ImageNet pre-trained models and +3.9\%, +2.4\% and +1.2\% mIoU improvement compared to vanilla fine-tuning of a CLIP models based on ResNet-50, ResNet-101~\cite{he2016deep} and ViT-B~\cite{vit} respectively. We also observe significant improvements in object detection and instance segmentation tasks. Notably, we show a ResNet-101 model equipped with our method and a lightweight semantic FPN decoder can achieve 46.5\% mIoU on ADE20K, which outperforms state-of-the-art solutions like DeepLabV3+~\cite{deeplabv3+} and UperNet~\cite{upernet} with only 1/3 computation.  

Moreover, our framework can also be applied to \emph{any} backbone models by using the pre-trained language model to guide the training of dense prediction tasks. We observe significant improvements by applying DenseCLIP to ImageNet pre-trained ResNets~\cite{he2016deep} and recent Swin Transformers~\cite{swin} with slight computation overhead. We expect our method to be a new and generic paradigm to improve dense prediction models with guidance from pre-trained language models.

\section{Related Work}
\paragrapha[0pt]{Pre-training and fine-tuning.}
The revolution of computer vision in the past decade has been driven by the ``pre-training + fine-tuning'' paradigm. Specifically, it first pre-trains models on large-scale datasets (\textit{e.g.,} ImageNet~\cite{deng2009imagenet}, JFT~\cite{DBLP:conf/iccv/SunSSG17}, Kinetics~\cite{carreira2017quo}, \textit{etc.}) in a supervised learning~\cite{he2016deep, vit, swin, rao2021global} or self-supervised learning manner~\cite{moco, simclr ,caron2021emerging, he2021masked}, and then fine-tunes the models on various downstream tasks. In NLP community, this framework has also been similarly and widely used ~\cite{GPT3} and recently evolves into a prompt paradigm~\cite{liu2021pre}, in which downstream tasks are reformulated to simulate the solved tasks in original pretraining process.
Inspired by these works, we explore to transfer the knowledge in large-scale vision-language pre-trained models to the downstream dense prediction tasks.

\paragrapha{Vision-language models.} 
There have been a series of works on the interaction of computer vision and natural language processing fields, \textit{e.g.,} text-to-image retrieval~\cite{CAMP}, image caption~\cite{DBLP:conf/icml/XuBKCCSZB15}, visual question answering~\cite{antol2015vqa}, referring segmentation~\cite{hu2016segmentation,LAVT,HINet} and so on. 
Among these works, vision-language pre-training has attracted growing attention during the past few years~\cite{ViLBERT, VL-BERT, clipbert}. As a milestone, Radford~\etal devise a large-scale pretraining model, named CLIP~\cite{clip}, which employs a contrastive learning strategy on a huge amount of image-text pairs, and shows impressive transferable ability over 30 classification datasets. Motivated by this work, a number of follow-ups have been proposed to improve the training strategy (\textit{e.g.,} CoOp~\cite{coop}, CLIP-Adapter~\cite{clip-adapter}, Tip-adapter~\cite{zhang2021tip}) or apply it to other domains (\textit{e.g.,} ActionCLIP~\cite{wang2021actionclip}). However, there are very few attempts on performing dense prediction tasks via the CLIP model. The work most related to ours is CPT~\cite{yao2021cpt}, which reformulates dense predictions into a fill-in-the-blank
problem by jointly marking co-referential parts of both image and text in color. Differently, we consider a standard dense prediction setting in this paper, where we use pixel-text relationships to guide the training of dense prediction models and further optimize language embedding with image context with a context-aware prompting method.

\paragrapha{Dense prediction.} Compared with conventional instance-level classification problem, dense prediction tasks (\textit{e.g.,} semantic segmentation~\cite{ade}, instance segmentation~\cite{coco}, object detection~\cite{coco}) are more challenging as they requires to model the finer-grained representation at the pixel level or region level. 
Following the ``pre-training + fine-tuning'' paradigm, previous literatures have developed various dense prediction models like FCN~\cite{fcn}, PSPNet~\cite{pspnet}, FPN~\cite{fpn}, UperNet~\cite{upernet}, and many others. 
To alleviate the heavy annotation cost in previous supervised pre-training settings, a number of self-supervised pre-training approaches have been proposed for dense prediction~\cite{wang2021dense, xie2021propagate, xie2021self, caron2021emerging}. Orthogonal to these prior arts, we introduce a new fine-tuning strategy that leverages the knowledge in the large-scale vision-language pre-trained model and uses the language information to guide the learning process.

\section{Approach}

\begin{figure*}[t]
  \centering
  \includegraphics[width=0.9\linewidth]{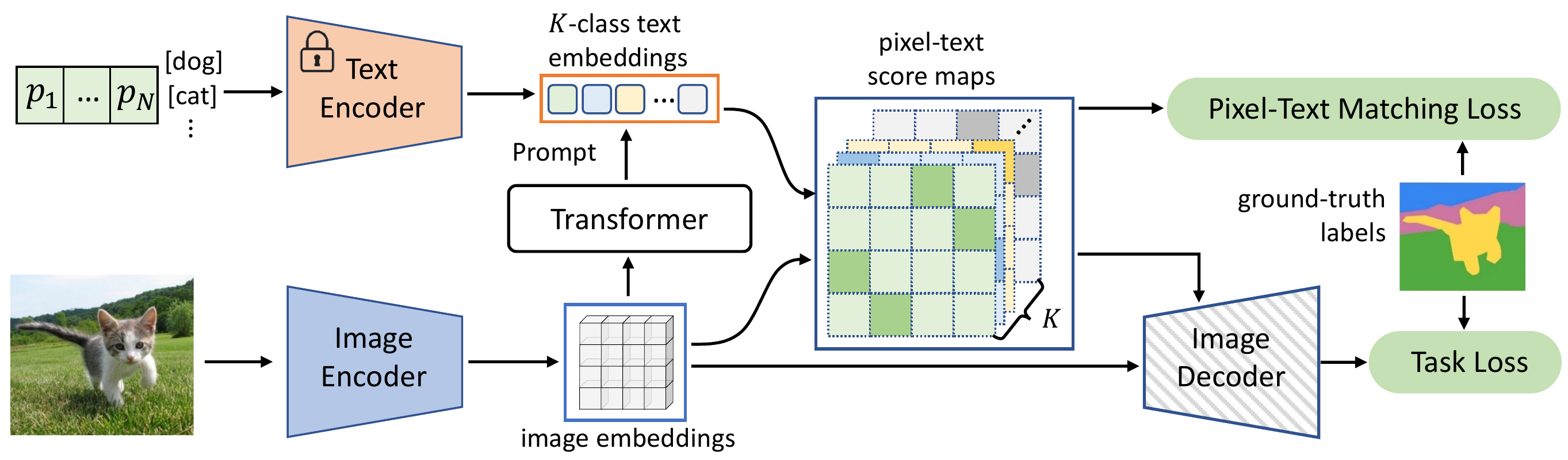}
   \caption{The overall framework of DenseCLIP. DenseCLIP first extracts the image embeddings and $K$-class text embeddings, and then calculates pixel-text score maps to convert the original image-text matching problem in CLIP to pixel-text matching for dense prediction. 
   These score maps are fed into decoder and also supervised using the ground-truth labels. To better exploit the pre-trained knowledge, DenseCLIP uses the contextual information in images to prompt the language model with a Transformer module. } \vspace{-10pt}
   \label{fig:framework}
\end{figure*}

\subsection{Preliminaries: Overview of CLIP}

We begin by reviewing the Contrastive Language-Image Pre-training (CLIP)~\cite{clip} framework to illustrate the motivation of our method. CLIP consists of two encoders, including an image encoder (ResNet~\cite{he2016deep} or ViT~\cite{vit}) and a text encoder (Transformer~\cite{transformer}). The goal of CLIP is to align the embedding spaces of visual and language during pre-training through a contrastive objective. 

To learn more transferable
pre-trained knowledge, CLIP collects 400 million
image-text pairs for model training. To transfer knowledge of CLIP to downstream classification task, a simple yet effective way~\cite{clip} is to construct a set of text prompts based on a template such as ``\texttt{a photo of a [CLS]}.'', where \texttt{[CLS]} can be replaced by the actual class names. Then given an image, one can use CLIP to compute the similarities between the image and the text prompts in the embedding space and the class with the highest score is regarded as the final prediction. Recently, several works~\cite{coop,clip-adapter} have shown CLIP can 
obtain strong classification performance with few examples. Therefore, it raises an interesting question: \emph{whether the impressive ability of CLIP can be transferred to more complex vision tasks like dense prediction?}

However, the extension is nontrivial. Firstly, how to leverage the visual-language pre-trained model in dense prediction tasks is a barely visited question. Although a simple solution is to only use the image encoder like a pre-trained 2D backbone, we argue that the language priors contained in the text encoder are also of great importance. Secondly, unlike the classification considered in~\cite{coop,clip-adapter}, 
transferring the knowledge from CLIP to dense prediction is more difficult due to the substantial gap between the upstream contrastive pre-training task and the downstream per-pixel prediction task, where the former considers instance-level representation of both images and texts, and the latter is only based on the visual information but expects pixel-level outputs.

\subsection{Language-Guided Dense Prediction}
\label{sec:dense_prediction}

To solve the above issues, we propose our language-guided dense prediction framework, which can better leverage the language priors in CLIP pre-trained models. The pipeline of our framework is shown in Figure~\ref{fig:framework}. One of our important findings is that apart from the global image feature, we can also extract a language-compatible feature map from the last layer of the CLIP image encoder. To show this, we start by describing the architecture of the CLIP image encoder in detail. Take the ResNet~\cite{he2016deep} encoder for example, there are 4 stages in total and we denote the feature maps as $\{\mathbf{x}_i\}_{i=1}^4$. Different from the original ResNet~\cite{he2016deep}, CLIP makes a small modification~\cite{clip} by adding an attention pooling layer. Specifically, CLIP first performs global average pooling to $\mathbf{x}_4\in\mathbb{R}^{H_4W_4\times C}$ to obtain a global feature $\bar{\mathbf{x}}_4\in\mathbb{R}^{1\times C}$, where $H_4, W_4, C$ are the height, width and the number of channels of the feature maps from the 4-th stage of the backbone. The concatenated features $[\bar{\mathbf{x}}_4, \mathbf{x}_4]$ are then fed into an multi-head self-attention layer~\cite{transformer} (MHSA):
\begin{equation}
    [\bar{\mathbf{z}}, \mathbf{z}] = \mhsa([\bar{\mathbf{x}}_4, \mathbf{x}_4]).
\end{equation}
In the standard training process of CLIP, the global feature $\bar{\mathbf{z}}$ is used as the output of the image encoder while the other outputs $\mathbf{z}$ are usually neglected. However, we find $\mathbf{z}$ has two interesting properties: (1) $\mathbf{z}$ still retains sufficient spatial information thus can serve as a feature map. (2) since the MHSA is symmetric to each input element, $\mathbf{z}$ might behave similarly to $\bar{\mathbf{z}}$, which aligns well with the language features. Based on the above observations, we can use $\mathbf{z}$ as a language-compatible feature map. It is also noted that for architectures like ViT~\cite{vit}, $\mathbf{z}$ can be obtained similarly by excluding the class token of outputs. 

To obtain the text features, we can construct text prompts from the template ``\texttt{a photo of a [CLS]}.'' with $K$ class names, and the use CLIP text encoder to extract the features as $\mathbf{t}\in \mathbb{R}^{K\times C}$. We then compute the pixel-text score maps using the language-compatible feature map $\mathbf{z}$ and the text features $\mathbf{t}$ by:
\begin{equation}
    \mathbf{s} = \hat{\mathbf{z}}{\hat{\mathbf{t}}}^\top, \quad \mathbf{s}\in\mathbb{R}^{H_4W_4\times K},
\end{equation}
where $\hat{\mathbf{z}}$ and $\hat{\mathbf{t}}$ are the $\ell_2$ normalized version of $\mathbf{z}$ and $\mathbf{t}$ along the channel dimension. The score maps characterize the results of pixel-text matching, which is one of the most crucial ingredients in our framework. Firstly, the score maps can be viewed as segmentation results with a lower resolution, and thus we can use them to compute an auxiliary segmentation loss. Secondly, we can concatenate the score maps to the last feature map to explicitly incorporate language priors, \ie, $\mathbf{x}_4' = [\mathbf{x}_4, \mathbf{s}]\in \mathbb{R}^{H_4W_4\times (C+K)}$. Our framework is model-agnostic because the modified feature maps can be directly used as usual in segmentation or detection with some minor modifications (\eg, the input dimension of FPN~\cite{fpn}).

\subsection{Context-Aware Prompting}
Previous efforts~\cite{coop,clip-adapter} have already proved that mitigating the domain gaps in visual or language can significantly improve the performance of CLIP models on downstream tasks. Therefore, instead of using the vanilla human pre-defined templates, we seek for other methods to improve the text features $\mathbf{t}$.

\paragrapha{Language-domain prompting.} Different from the original CLIP that uses human-designed templates like ``\texttt{a photo of a [CLS]}.'' as text prompts, CoOp~\cite{coop} introduces learnable textual contexts to achieve better transferability in downstream classification tasks by directly optimizing the contexts using back-propagation. Inspired by CoOp~\cite{coop}, we also use learnable textual contexts in our framework as a baseline, which only includes language-domain prompting. The input of the text encoder then becomes:
\begin{equation}
    [\mathbf{p}, \mathbf{e}_k], \quad 1\le k\le K,
    \label{equ:langauge_prompting}
\end{equation}
where $\mathbf{p} \in\mathbb{R}^{N\times C}$ are the learnable textual contexts and $\mathbf{e}_k\in\mathbb{R}^C$ is the embedding for the name of the $k$-th class.

\paragrapha{Vision-to-language prompting.} Including descriptions of visual contexts can make the text more accurate. For example, ``\texttt{a photo of a cat in the grass}.'' is more accurate than ``\texttt{a photo of a cat}.''. Therefore, we investigate how to use visual contexts to refine the text features. Generally, we can use the cross-attention mechanism in Transformer decoder~\cite{transformer} to model the interactions between vision and language.

\begin{figure}[t]
  \centering
  \includegraphics[width=0.99\linewidth]{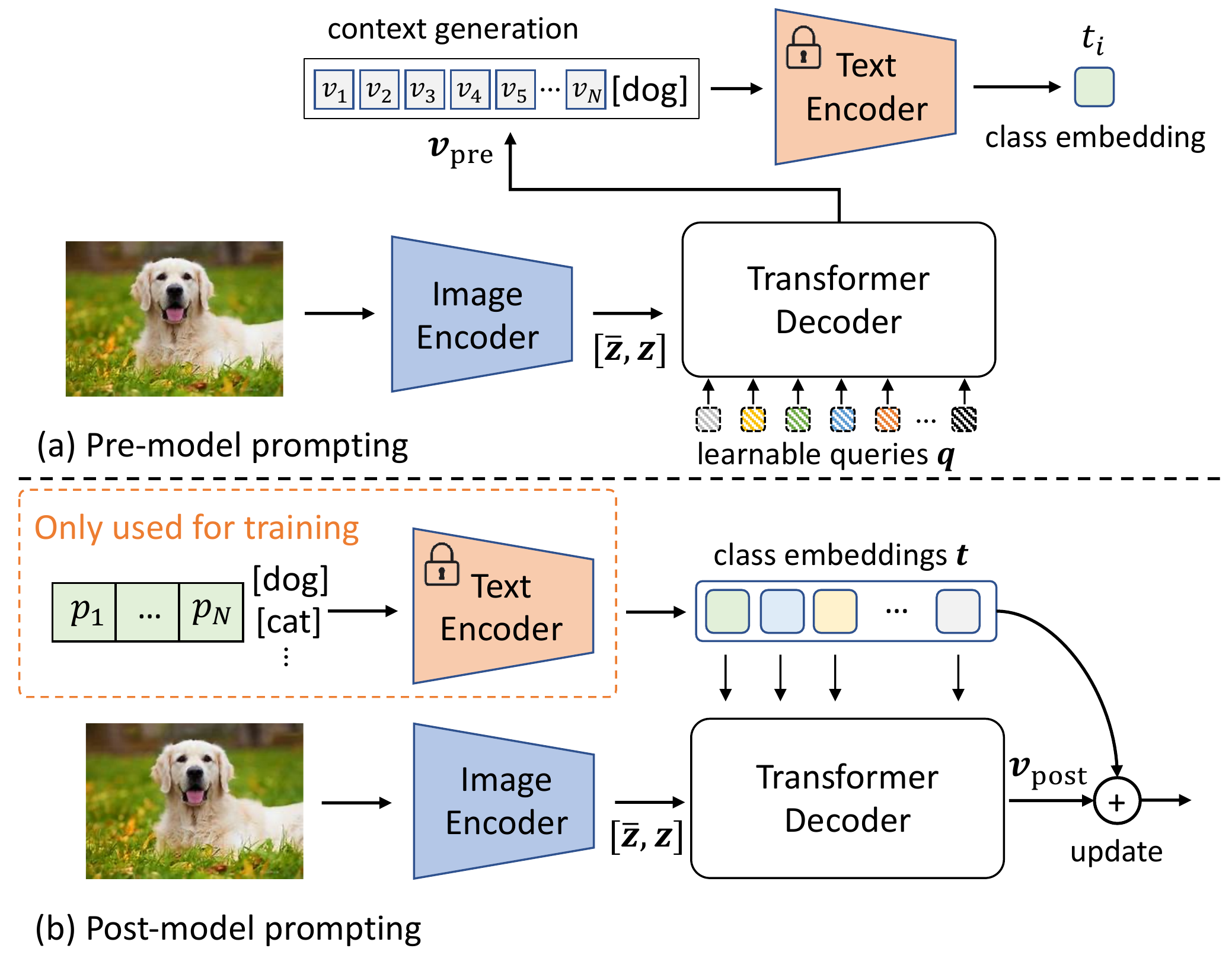}
   \caption{\textbf{Two different strategies of context-aware prompting.} The pre-model prompting directly uses the image contexts to generate the desired text inputs, while post-model prompting refines the class embedding instead.} \vspace{-10pt}
   \label{fig:prompting}
\end{figure}

We propose two different strategies of context-aware prompting, which is shown in Figure~\ref{fig:prompting}.
The first strategy we consider is the pre-language-model prompting, or \textit{pre-model prompting} for short. We pass the features $[\bar{\mathbf{z}}, \mathbf{z}]$ to a Transformer decoder to encode visual contexts:
\begin{equation}
    \mathbf{v}_{\rm pre} = \transdecoder(\mathbf{q}, [\bar{\mathbf{z}}, \mathbf{z}]),
\end{equation}
where $\mathbf{q}\in\mathbb{R}^{N\times C}$ are a set of learnable queries and $\mathbf{v}_{\rm pre}\in \mathbb{R}^{N\times C}$ are the extracted visual contexts. We replace the $\mathbf{p}$ in Equation~\eqref{equ:langauge_prompting} by the visual contexts $\mathbf{v}$ to form the input of the text encoder. Since the input of the text encoder is modified, we refer to this version as \textit{pre-model prompting}.

Another choice is to refine the text features after the text encoder, namely \textit{post-model prompting}. In this variant, we use CoOp~\cite{coop} to generate text features and directly use them as the queries of the Transformer decoder:
\begin{equation}
     \mathbf{v}_{\rm post} = \transdecoder(\mathbf{t}, [\bar{\mathbf{z}}, \mathbf{z}]).
\end{equation}
This implementation encourage the text features to find most related visual clues. We then update the text features through a residual connection:
\begin{equation}
    \mathbf{t} \leftarrow \mathbf{t} + \bm{\gamma}\mathbf{v}_{\rm post},
\end{equation}
where $\bm{\gamma}\in \mathbb{R}^C$ is a learnable parameter to control the scaling of the residual. $\bm{\gamma}$ is initialized with very small values (\eg, $10^{-4}$) to maximally preserve the language priors from the text features.

Although the two variants target the same goal, we prefer the post-model prompting for mainly two reasons: (1) The post-model prompting is efficient. The pre-model prompting requires extra forward passes of the text encoder during inference since its input is dependent on the image. In the case of post-model prompting, we can store the extracted text features after training and thus can reduce the overhead brought by the text encoder during inference. (2) Our empirical results show the post-model prompting can achieve better performance than pre-model prompting.

\subsection{Instantiations}

\noindent \textbf{Semantic segmentation. } As discussed in Section~\ref{sec:dense_prediction}, our framework is model-agnostic and can be applied to any dense prediction pipelines. Moreover, we propose to use an auxiliary objective to make better use of our pixel-text score maps in segmentation. Since the score maps $\mathbf{s}\in \mathbb{R}^{H_4W_4\times K}$ can be viewed as smaller segmentation results, we therefore compute a segmentation loss on it:
\begin{equation}
    \mathcal{L}_{\rm aux}^{\rm seg} = \ce(\softmax(\mathbf{s}/\tau), \mathbf{y}),
\end{equation}
where $\tau=0.07$ is a temperature coefficient following~\cite{moco} and $\mathbf{y}\in \{1, \ldots, K\}^{H_4W_4}$ is the ground truth label. The auxiliary segmentation loss can help the feature map to recover its locality faster, which is beneficial to dense prediction tasks for both segmentation and detection.

\paragrapha{Object detection \& instance segmentation. } In this case, we do not have ground truth segmentation labels. To construct a similar auxiliary loss as in segmentation, we use the bounding box and the label to build a binary target $\tilde{\mathbf{y}}\in \{0, 1\}^{H_4W_4\times K}$. The auxiliary objective can be defined as a binary cross-entropy loss:
\begin{equation}
    \mathcal{L}_{\rm aux}^{\rm det} = \bce(\sigmoid(\mathbf{s}/\tau), \tilde{\mathbf{y}}).
\end{equation}

\paragrapha{Applications to \emph{any} backbone models. } Another interesting usage of our framework is that we can replace the image encoder of CLIP with \textit{any} backbones (\eg, ImageNet pre-trained models and self-supervised models). Although there might be no strong relation between the outputs of the visual backbone and the text encoder, the backbone can learn better and faster with language guidance. In other words, we can leverage the language priors from the pre-trained text encoder to improve the performance of any pre-trained image backbone, which makes DenseCLIP a more generic framework to improve dense prediction with the natural language priors learned from large-scale pre-training.

\section{Experiments}

\begin{table*}[t]
\centering
 \caption{\textbf{Semantic segmentation results on ADE20K}. We compare the performance of DenseCLIP and existing methods when using the same backbone. We report the mIoU of both single-scale and multi-scale testing, the FLOPs and the number of parameters. The FLOPs are measured with $1024\times1024$ input using the \texttt{fvcore} library. The results show that our DenseCLIP outperforms other methods by large margins with much lower complexity. Our models and our baselines that are trained using identical settings are highlighted in gray.}
 \label{tab:segmentation}  \vspace{-5pt}
 \centering
 \small \renewcommand{\arraystretch}{0.67}
 \setlength{\tabcolsep}{10pt}
\begin{tabular}{l>{\columncolor{white}[3pt][\tabcolsep]}llrrr>{\columncolor{white}[\tabcolsep][3pt]}r}
\toprule
Backbone                   & Method  & Pre-train  & mIoU (SS) & mIoU (MS) & GFLOPs  & Params (M)\\ \midrule
\multirow{11}{*}{ResNet-50}  
                    & FCN~\cite{fcn} & ImageNet  & 36.1    & 38.1   & 793.3   & 49.6    \\
              & EncNet~\cite{encnet} & ImageNet   & 40.1    & 41.7   & 565.6   & 36.1   \\
              & PSPNet~\cite{pspnet} & ImageNet   & 41.1    & 41.9   & 716.2   & 49.1   \\
               & CCNet~\cite{ccnet}  & ImageNet   & 42.1    & 43.1   & 804.0   & 49.9   \\
     & DeeplabV3+~\cite{deeplabv3+}  & ImageNet   & 42.7    & 43.8   & 711.5   & 43.7   \\
           & UperNet~\cite{upernet}  & ImageNet   & 42.1    & 42.8   & 953.2   & 66.5   \\
           & DNL~\cite{yin2020dnl} & ImageNet & 41.9 & 43.0 & 939.3 & 50.1 \\
                            &  \cellcolor{Gray}Semantic FPN~\cite{semfpn}   & \cellcolor{Gray}ImageNet   & \cellcolor{Gray}38.6    & \cellcolor{Gray}40.6   & \cellcolor{Gray}227.1 & \cellcolor{Gray}31.0  \\
                        & \cellcolor{Gray}CLIP + Semantic FPN  &\cellcolor{Gray}CLIP  & \cellcolor{Gray}39.6    & \cellcolor{Gray}41.6   & \cellcolor{Gray}248.8   & \cellcolor{Gray}31.0  \\
                 &  \cellcolor{Gray}DenseCLIP + Semantic FPN & \cellcolor{Gray}CLIP  & \cellcolor{Gray}\textbf{43.5}    & \cellcolor{Gray}\textbf{44.7}      & \cellcolor{Gray}269.2   & \cellcolor{Gray}50.3 \\ \midrule
\multirow{12}{*}{ResNet-101} 
                    & FCN~\cite{fcn} & ImageNet   & 39.9    & 41.4   & 1104.4   & 68.6  \\
              & EncNet~\cite{encnet} & ImageNet   & 42.6    & 44.7   & 876.8   & 55.1   \\
              & PSPNet~\cite{pspnet}  & ImageNet  & 43.6    & 44.4   &  1027.4  & 68.1  \\
               & CCNet~\cite{ccnet}   & ImageNet  & 44.0    & 45.2   &  1115.2  & 68.9  \\
     & DeeplabV3+~\cite{deeplabv3+}  & ImageNet   & 44.6    & 46.1   &  1022.7  & 62.7  \\
           & UperNet~\cite{upernet}   & ImageNet  & 43.8    & 44.8   & 1031.0   & 85.5    \\
           & OCRNet~\cite{YuanCW20ocr} & ImageNet & 45.3 & - & 923.9 & 55.5 \\
           & DNL~\cite{yin2020dnl} & ImageNet & 44.3 & 45.8 & 1250.5 & 69.1\\
                            & \cellcolor{Gray}Semantic FPN~\cite{semfpn}   & \cellcolor{Gray}ImageNet   & \cellcolor{Gray}40.4    &     \cellcolor{Gray}42.3   & \cellcolor{Gray}304.9   & \cellcolor{Gray}50.0  \\
                         & \cellcolor{Gray}CLIP + Semantic FPN & \cellcolor{Gray}CLIP   & \cellcolor{Gray}42.7    &   \cellcolor{Gray}44.3     & \cellcolor{Gray}326.6   & \cellcolor{Gray}50.0 \\
                  & \cellcolor{Gray}DenseCLIP + Semantic FPN & \cellcolor{Gray}CLIP  & \cellcolor{Gray}\textbf{45.1}    & \cellcolor{Gray}\textbf{46.5}   & \cellcolor{Gray}346.3   & \cellcolor{Gray}67.8  \\ \midrule
\multirow{5}{*}{ViT-B} 
                       & SETR-MLA-DeiT~\cite{SETR}  & ImageNet         & 46.2    & 47.7   &  - &  -\\
                       & \cellcolor{Gray}Semantic FPN~\cite{semfpn}           & \cellcolor{Gray}ImageNet   &   \cellcolor{Gray}48.3  & \cellcolor{Gray}50.9 & \cellcolor{Gray}1037.4  &     \cellcolor{Gray}100.8\\
                       & \cellcolor{Gray}Semantic FPN~\cite{semfpn}       & \cellcolor{Gray}ImageNet-21K      &   \cellcolor{Gray}49.1 & \cellcolor{Gray}50.4 & \cellcolor{Gray}1037.4  &    \cellcolor{Gray}100.8 \\
                       & \cellcolor{Gray}CLIP + Semantic FPN    & \cellcolor{Gray}CLIP    & \cellcolor{Gray}49.4    &   \cellcolor{Gray}50.3     & \cellcolor{Gray}1037.4    & \cellcolor{Gray}100.8 \\
                     & \cellcolor{Gray}DenseCLIP + Semantic FPN  & \cellcolor{Gray}CLIP  & \cellcolor{Gray}\textbf{50.6}    &    \cellcolor{Gray}\textbf{51.3}   & \cellcolor{Gray}1043.1    & \cellcolor{Gray}105.3  \\ \bottomrule
\end{tabular} \vspace{-10pt}
\end{table*}

\begin{table}[t]
  \centering
  \caption{\textbf{Ablation study.} We demonstrate that performing post-model vision-to-language prompting can yield the better performance with fewer extra FLOPs and parameters.} \vspace{-5pt}
  \small
  \setlength{\tabcolsep}{2pt}
    \begin{tabular}{lccclrr}\toprule
    \multicolumn{1}{c}{\multirow{2}[0]{*}{Pre-train}} & \multirow{2}[0]{*}{
    \makecell[c]{Language \\Prompt}} & \multicolumn{2}{c}{V$\rightarrow$L Prompt} & \multirow{2}[0]{*}{
    \makecell[c]{mIoU \\ (\%)}} & \multirow{2}[0]{*}{
    \makecell[c]{FLOPs\\(G)}}& \multirow{2}[0]{*}{
    \makecell[c]{Params\\(M)}}  \\\cmidrule{3-4}
          &       & pre & post &      &    & \\\midrule
    ImageNet &    &       &       & 38.6  & 227  & 31.0  \\\midrule
    CLIP  &       &       &       & 39.6$_{\cb{(+1.0)}}$  & 249  & 31.0  \\
    CLIP  & \cmark  &     &       & 42.1$_{\cb{(+3.5)}}$  & 269  & 46.5  \\
    CLIP  & \cmark     & \cmark &       & 42.9$_{\cb{(+4.3)}}$  & 368  & 116.9  \\
    CLIP  & \cmark     &       & \cmark     & \textbf{43.5}$_{\cb{(+4.9)}}$  & 269  & 50.2  \\\bottomrule
    \end{tabular} \vspace{-12pt}
  \label{tab:ablation}%
\end{table}%

To evaluate the effectiveness of our DenseCLIP, we conduct extensive experiments on dense prediction tasks including semantic segmentation, object detection and instance segmentation. The following subsections describe the details of the experiments, results and analyses.

\subsection{Semantic Segmentation}

\paragrapha{Setups.} We start by evaluating our DenseCLIP on ADE20K~\cite{ade}, a challenging large-scale semantic segmentation dataset that covers a broad range of 150 categories. ADE20K  contains 20K images for training and 2K images for validation. Following common practice~\cite{ccnet,upernet}, we report the mIoU on the validation set. For fair comparisons, we also include the FLOPs and the number of parameters.

\paragrapha{Implementation details.} We experiment with the popular Semantic FPN~\cite{semfpn} framework to evaluate our DenseCLIP. Specifically, we apply the pre-trained image encoder of the CLIP as the segmentation backbone, and directly use the Semantic FPN~\cite{semfpn} as the decoder. We consider three kinds of image backbones including ResNet-50~\cite{he2016deep}, ResNet-101~\cite{he2016deep}, and ViT-B~\cite{vit}. For language-domain prompting, we use a context length of 8. The Transformer decoder to extract visual contexts consists of 6 layers and we set the number of heads as 4. We fix the text encoder during training to preserve the natural language knowledge learned from large-scale pre-training. To reduce the computational costs, we project both the image embeddings and the text embeddings to a lower dim (256) before the Transformer module.  We empirically find that directly fine-tuning CLIP models to dense prediction with the default training strategies in~\cite{mmsegmentation} will lead to unsatisfactory results (only 21.9\% mIoU on ADE20K, which is 15.6\% lower than its ImageNet pre-trained counterpart). Therefore, two key modifications are made compared to the default configurations: (1) we use AdamW~\cite{adamw} instead of the default SGD inspired by recent progress in vision Transformers~\cite{deit,pvt,swin}; (2) to better preserve the pre-trained weights, we set the learning rate of the image encoder as $1/10$ of the other parameters. We also adopt the above training strategies to our baselines in ablation studies for fair comparisons (+1.1\% mIoU over the ImageNet pre-trained ResNet-50 with the default settings in~\cite{mmsegmentation}).
 
 \paragrapha{Main results.} We report the semantic segmentation results of our DenseCLIP with three different backbones on ADE20K in Table~\ref{tab:segmentation}. We include the FLOPs, the number of parameters, and the mIoU in both single-scale (SS) and multi-scale (MS) testings. The experiments results show that for the same backbone, our DenseCLIP with a simple Semantic FPN can outperform the state-of-the-art methods that use more sophisticated decoders by large margins. Unlike previous works that use dilated backbones (ResNet-D8~\cite{dilated,ccnet,pspnet,YuanCW20ocr}), the ResNet encoder in DenseCLIP is more close to standard ResNet thus our DenseCLIP has much fewer FLOPs. Besides, our DenseCLIP is +4.9\%, +4.7\%, and +2.3\% mIoU (SS) higher than the original ImageNet pre-trained baselines on ResNet-50, ResNet-101 and ViT-B backbones with acceptable extra computation cost. DenseCLIP is also +3.9\%, +2.4\%, and +1.2\% mIoU higher than the vanilla fine-tuning strategy (CLIP + Semantic FPN).

\paragrapha{Ablation studies.} To further demonstrate the effects of different components of our DenseCLIP, we perform detailed ablation studies with the ResNet-50~\cite{he2016deep} backbone and the results are shown in Table~\ref{tab:ablation}. Firstly, we show by adopting a better training strategy aforementioned the ResNet-50 baseline we implemented has a higher mIoU than~\cite{mmsegmentation} (38.6\% \vs 37.5\%). Secondly, we find that CLIP pre-trained ResNet-50 outperforms the ImageNet pre-trained one by 1\%, which indicates that large-scale vision language pre-trained model can be better transferred to downstream vision tasks. To better leverage the language priors, we adopt our language-guided with language-domain prompt and witness a significant performance boost (+2.5\% mIoU). Finally, we compare the two methods to perform vision-language prompting to incorporate visual contexts. We find both the pre-model and post-model prompting can improve the performance, while the post-model prompting is better and more computationally efficient. Therefore, we choose the post-model prompting as the default configuration in all the rest experiments.

\paragrapha{Effects of language-guided pre-training and fine-tuning.} We compare the performance on ADE20K of different pre-training and fine-tuning strategies to better reveal the potential of language-guided paradigm, which is shown in Figure~\ref{fig:pre-trains}.  We consider supervised pre-training on ImageNet1K~\cite{deng2009imagenet} and ImageNet21K~\cite{deng2009imagenet,ridnik2021imagenet}, self-supervised pre-training via MoCoV2~\cite{moco} and DenseCL~\cite{wang2021dense}, and the vision-language pre-training. We show that the vision-language pre-trained model (CLIP) can outperform ImageNet1K pre-trained model by vanilla fine-tuning. Furthermore, through the language-guided fine-tuning with context-aware prompting, our DenseCLIP surpasses even the ImageNet21K pre-trained model. These promising results demonstrate that language-priors can largely facilitate vision models in downstream dense prediction tasks.

\subsection{Object Detection and Instance Segmentation}
\paragrapha{Setups.} We also conduct experiments to apply our DenseCLIP to object detection and instance segmentation tasks on COCO~\cite{coco}, which contains 118K training images and 5K validation images. We adopt two widely used frameworks, RetinaNet~\cite{focal} and Mask R-CNN~\cite{he2017mask}. Following~\cite{he2017mask}, we report the standard AP, AP at IoU=0.5/0.75, and cross-scale AP. For Mask R-CNN, we report both the mAPs for object detection and instance segmentation since these two tasks are performed simultaneously. 

\paragrapha{Implementation details.} For object detection, we adopt ResNet-50 and ResNet-101 as backbones. We train all the models for 12 epochs using AdamW optimizer with batch size 16 as in~\cite{mmdetection}. Specifically for RetinaNet, we witness a super large loss at the start of training that makes the model hard to converge. Therefore, we use gradient clipping with a max $\ell_2$ norm of 0.1 to protect the pre-trained weights.

\paragrapha{Results analysis.}
The results using the RetinaNet~\cite{focal} and the Mask R-CNN~\cite{he2017mask} are summarized in Table~\ref{tab:retina} and Table~\ref{tab:mask_rcnn}, respectively. For object detection with RetinaNet, we compare DenseCLIP with ImageNet1K pretrained model and vanilla CLIP fine-tuning on detection task. One can observe that DenseCLIP outperforms the ImageNet1K pretrained model by +1.5\% and +2.6\% AP. Meanwhile, it also improves the vanilla fine-tuning strategy by +0.9\% and +0.6\% AP on both ResNet-50 and ResNet-101 backbones.

For Mask R-CNN, we observe that DenseCLIP achieves consistent improvement on both object detection and instance segmentation tasks within an affordable computational budget. Especially for instance segmentation, our DenseCLIP outperforms the ImageNet1K pre-trained model with +2.9\% and +2.5\% mask AP on both ResNet50 and ResNet101 backbones and also outperforms the vanilla fine-tuning strategy with +0.8\% and +0.7\% mask AP. The significant improvements of DenseCLIP on the instance segmentation task suggest that our pixel-text matching is conceptually suitable for segmentation.

\begin{table}[t]
  \centering
  \caption{\textbf{Object detection on COCO val2017 using RetinaNet~\cite{focal} framework.} We compare our DenseCLIP framework to the vanilla fine-tuning of ImageNet/CLIP pre-trained models. We find DenseCLIP can better make use of the language priors to facilitate better training.}
  \small
  \adjustbox{width=\linewidth}{
  \setlength{\tabcolsep}{2pt}
    \begin{tabular}{lcccccccc}\toprule
    Model & \makecell{FLOPs\\(G)} & \makecell{Params\\(M)}& AP & AP$_{50}$ & AP$_{75}$ & AP$_S$ & AP$_M$ & AP$_L$ \\\toprule
    RN50-IN1K~\cite{he2016deep} & 239    & 38   & 36.3  & 55.3  & 38.6  & 19.3  & 40.0 & 48.8 \\
    RN50-CLIP~\cite{clip} & 265 & 38 & 36.9  & 57.7  & 39.1  & 22.5  & 40.7  & 47.1 \\
    RN50-DenseCLIP & 285 & 60 & \textbf{37.8}  & \textbf{59.9}  & \textbf{40.0}    & \textbf{24.8}  & \textbf{42.0 }   & \textbf{47.9} \\\midrule
    RN101-IN1K~\cite{he2016deep} &    315 & 57      & 38.5  & 57.6  & 41.0    & 21.7  & 42.8  & 50.4 \\
    RN101-CLIP~\cite{clip} &    341 & 57     & 40.5  & 61.6  & 43.4  & 25.6  & 44.6  & 51.3 \\
    RN101-DenseCLIP &   360 & 78    & \textbf{41.1}  & \textbf{63.4}  & \textbf{44.1}  & \textbf{26.9}  & \textbf{45.5}  & \textbf{52.4} \\\bottomrule
    \end{tabular}%
}\vspace{-10pt}
  \label{tab:retina}%
\end{table}%

\begin{table*}[t]
  \centering
  \caption{\textbf{Object detection and instance segmentation results on COCO val2017 using Mask R-CNN~\cite{he2017mask} framework.} Our DenseCLIP outperforms ImageNet/CLIP pre-trained baseline models, especially on the instance segmentation task.}
  \renewcommand{\arraystretch}{0.8}
  \setlength{\tabcolsep}{3.5pt}
    \begin{tabular}{lcc|cccccc|cccccc}\toprule
    Model & \makecell{FLOPs\\(G)} & \makecell{Params\\(M)} & AP$^{\rm b}$ & AP$^{\rm b}_{50}$ & AP$^{\rm b}_{75}$ & AP$^{\rm b}_{S}$ & AP$^{\rm b}_{M}$ & AP$^{\rm b}_{L}$ & AP$^{\rm m}$  & AP$^{\rm m}_{50}$ & AP$^{\rm m}_{75}$ & AP$^{\rm m}_{S}$ & AP$^{\rm m}_{M}$ & AP$^{\rm m}_{L}$ \\\midrule
    RN50-IN1K~\cite{he2016deep} &   275 & 44    & 38.2  & 58.8  & 41.4  & 21.9  & 40.9  & 49.5  & 34.7  & 55.7  & 37.2  & 18.3  & 37.4  & 47.2  \\
    RN50-CLIP~\cite{clip} &  301 &  44    & 39.3  & 61.3  & 42.7  & 24.6  & 42.6  & 50.1  & 36.8  & 58.5  & 39.2  & 18.6  & 39.9  & 51.8  \\
    RN50-DenseCLIP &   327 &  67   & \textbf{40.2}  & \textbf{63.2}  & \textbf{43.9}  & \textbf{26.3}  & \textbf{44.2}  & \textbf{51.0}  & \textbf{37.6}  & \textbf{60.2}  & \textbf{39.8}  & \textbf{20.8}  & \textbf{40.7}  &\textbf{53.7}  \\\midrule
    RN101-IN1K~\cite{he2016deep} &   351 & 63    & 40.0  & 60.5  & 44.0  & 22.6  & 44.0  & 52.6  & 36.1  & 57.5  & 38.6  & 18.8  & 39.7  & 49.5  \\
    RN101-CLIP~\cite{clip} &   377 & 63   & 42.2  & 64.2  & \textbf{46.5}  & 26.4  & 46.1  & 54.0  & 38.9  & 61.4  & 41.8  & 20.5  & 42.3  & 55.1  \\
    RN101-DenseCLIP & 399 & 84 & \textbf{42.6}  & \textbf{65.1}  & \textbf{46.5}  & \textbf{27.7}  & \textbf{46.5}  & \textbf{54.2}  & \textbf{39.6}  & \textbf{62.4}  & \textbf{42.4}  & \textbf{21.4}  & \textbf{43.0}  & \textbf{56.2} \\\bottomrule
    \end{tabular}%
  \label{tab:mask_rcnn} \vspace{-10pt}
\end{table*}%

\subsection{DenseCLIP for Any Visual Backbone}
Previous experiments have demonstrated the effectiveness of our DenseCLIP framework. However, since DenseCLIP is specifically designed to leverage the visual-language relation contained in the pre-trained CLIP models, the generalization ability of DenseCLIP might be somehow doubted: \textit{Is DenseCLIP only suitable to CLIP image encoders?} To answer this question, we perform experiments to verify whether our DenseCLIP can also perform well with other backbones. The extension is actually straightforward: we can simply replace the CLIP image encoder with any given 2D pre-trained image model. Although there are no strong correlations between the feature maps of the new backbone and the text features output by the CLIP text encoder, we hypothesize that if we preserve the language priors by freezing the text encoder as before, the text encoder will guide the backbone to better adapt to downstream tasks. 

 To verify the above assumption, we choose two representative 2D models including ResNet~\cite{he2016deep}, the most widely used CNN model, and Swin~\cite{swin}, the recent state-of-the-art vision Transformer. Following the standard setting in~\cite{semfpn} and~\cite{swin}, we use the Semantic FPN~\cite{semfpn} framework for ResNet models and the UperNet~\cite{upernet} framework for Swin models. The experimental results are summarized in Table~\ref{tab:any}, where we report the mIoU on ADE20K of both the single-scale and multi-scale testing. We demonstrate that our DenseCLIP can consistently improve all the baseline models notably. Specifically, DenseCLIP can bring $\sim2.5\%$ single-scale mIoU improvement for ResNet-50/101 with semantic FPN~\cite{semfpn}, and $\sim0.8\%$ improvement for Swin-T/S with UperNet~\cite{upernet}. These results clearly show that our DenseCLIP can successfully guide any pre-trained 2D backbone by language priors to boost performance. Since the text encoder can be removed after training, our method provides a low-cost solution to improve arbitrary dense prediction models. Although these performances still lag behind our models with CLIP image encoders, the findings in this section provide a solution to generalize human knowledge learned from large-scale vision-language pre-training to a wider range of models. We expect this could be an interesting direction to connect vision and language researches in the future.

 \begin{figure}[t]
  \centering
  \includegraphics[width=\linewidth]{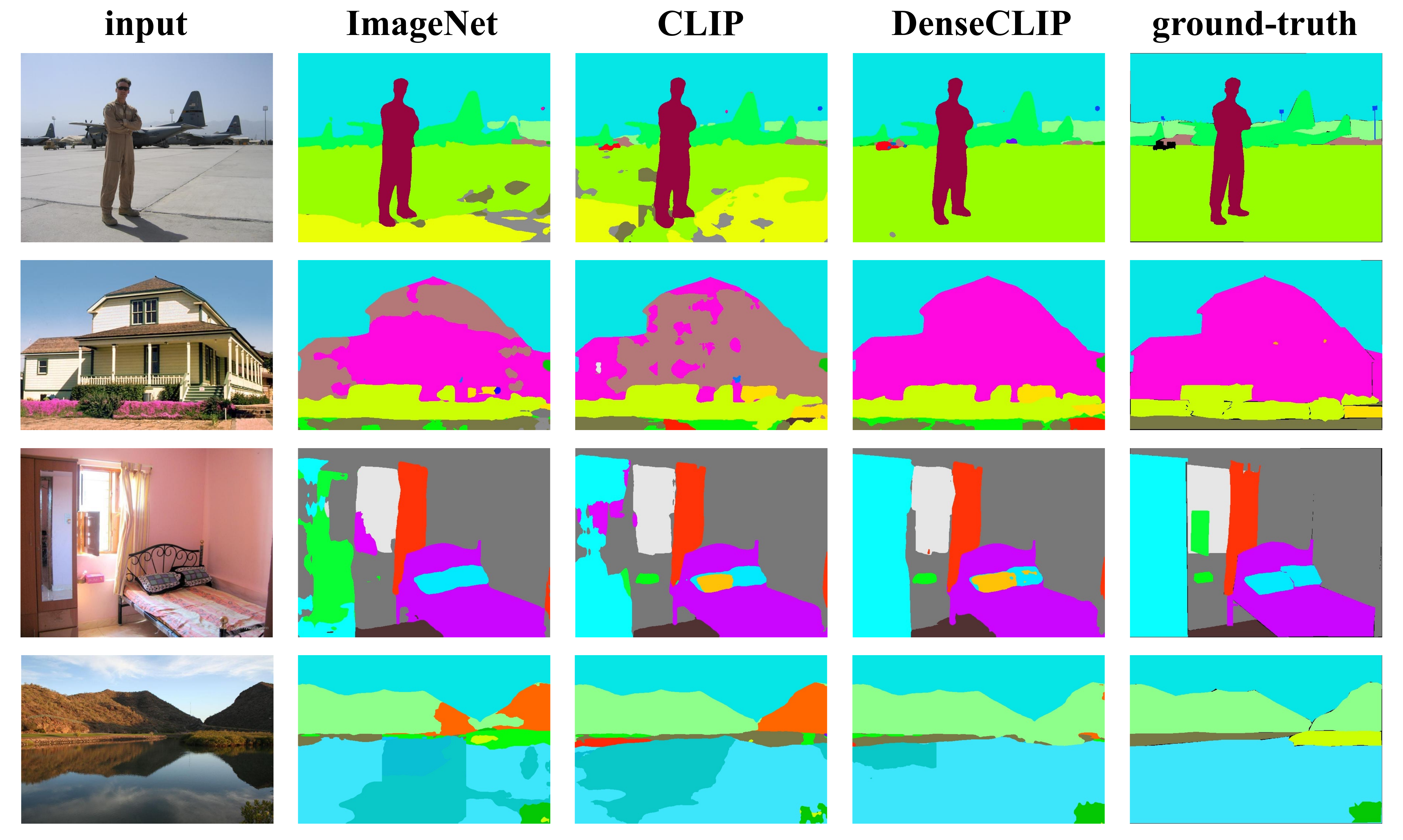}
   \caption{\textbf{Qualitative results on ADE20K.} We visualize the segmentation results on ADE20K validation set of our DenseCLIP based on ResNet-101 and two baseline models. }
   \label{fig:viz}
   \vspace{-10pt}
\end{figure}

\begin{table}[t]
  \centering
  \caption{\textbf{Applying DenseCLIP to \textit{any} backbone.} Image backbones (such as ImageNet pre-trained ResNet~\cite{he2016deep} and Swin~\cite{swin}) equipped with our DenseCLIP benefit from the language priors and enjoy significant performance boost. We report mIoU on ADE20K dataset for both single-scale (SS) and multi-scale (MS) testing.}
     \adjustbox{width=\linewidth}{
    \begin{tabular}{llll}\toprule
    \setlength{\tabcolsep}{2pt}
    Decoder & Method & \makecell[l]{mIoU (SS)\\(\%)} & \makecell[l]{mIoU (MS)\\(\%)} \\\midrule
    \multirow{4}[0]{*}{\makecell{Semantic \vspace{2pt}\\FPN~\cite{semfpn}}} & RN50~\cite{he2016deep} & 38.6  & 40.6  \\
          & RN50 + DenseCLIP & \textbf{41.0}$_{\cb{(+2.4)}}$  & \textbf{43.0}$_{\cb{(+2.4)}}$  \\\cmidrule{2-4}
          & RN101~\cite{he2016deep} & 40.4  & 42.3  \\
          & RN101 + DenseCLIP & \textbf{43.0}$_{\cb{(+2.6)}}$  & \textbf{45.2}$_{\cb{(+2.9)}}$  \\\midrule
    \multirow{4}[0]{*}{UperNet~\cite{upernet} } & Swin-T~\cite{swin} & 44.5  & 45.8  \\
          & Swin-T + DenseCLIP & \textbf{45.4}$_{\cb{(+0.9)}}$  & \textbf{46.5}$_{\cb{(+0.7)}}$  \\\cmidrule{2-4}
          & Swin-S~\cite{swin} & 47.6  & 49.5  \\
          & Swin-S + DenseCLIP & \textbf{48.3}$_{\cb{(+0.7)}}$  & \textbf{49.7}$_{\cb{(+0.2)}}$  \\\bottomrule
    \end{tabular}%
    }
  \label{tab:any}%
  \vspace{-10pt}
\end{table}%

\subsection{Visualization}
To better demonstrate the superiority of DenseCLIP, we provide several qualitative results in Figure~\ref{fig:viz}. We compare the segmentation maps of our method and the baselines and find DenseCLIP is better at identifying holistic objects.

\section{Conclusion and Discussion}
In this paper, we have presented a new framework, DenseCLIP, to transfer the knowledge from the vision-language pre-trained model (CLIP) to the downstream dense prediction tasks. 
DenseCLIP is a model-agnostic framework to use the pre-trained vision-language knowledge with the context-aware prompting strategy. The framework can be applied to various dense prediction tasks including semantic segmentation, object detection, and instance segmentation. We conducted extensive experiments to demonstrate the superior performance of our method.

\paragrapha{Limitations \& societal impact.} Although our method has achieved substantial improvement in segmentation, we find the improvements on detection are not such significant. We conjecture that it is because the pre-trained CLIP image encoder lacks locality since there is no such constraint during the pre-training of CLIP while object-centered tasks can only provide less dense supervision. We believe DenseCLIP can be further improved by introducing the dense supervision during pre-training or better recovering the locality after pre-training.  We develop a general method for dense prediction in this paper. Since our method is not for a specific application, it does not directly involve societal issues.

\subsection*{Acknowledgements}
This work was supported in part by the National Natural
Science Foundation of China under Grant 62125603, Grant U1813218, and in part
by a grant from the Beijing Academy of Artificial Intelligence
(BAAI).

{\small
\bibliographystyle{ieee_fullname}
\bibliography{ref}
}

\clearpage
\newpage
\appendix
\section*{Appendix: More Analysis}
We provide more analyses of both the design of our model and the training strategies in detail in the section.

\paragrapha{Effects of learning rate multipliers.} As discussed in Section 4.1, we found that the optimal learning rate for CLIP models and conventional ImageNet pre-trained models are different. Here we further investigate the effects of learning rate multiplier for image encoder and text encoder in Table~\ref{tab:lr}. We see both fixing the text encoder and using a lower learning rate for image encoder is beneficial to train the dense prediction model. Note that we observe a much lower performance ($<$30\% mIoU) when directly fine-tuning CLIP models with $1.0\times$ learning rate for the image encoder, which suggests our language guided method can largely stabilize the training process and make the final results less sensitive to the learning rate configuration.

\begin{table}[h]
  \centering
  \caption{Ablation study of the learning rate multiplier of text encoder and image encoder. We find freezing the text encoder and setting the lr multiplier of image encoder as 0.1 yields the best performance. The configuration used in our final models is highlighted in gray.} \small
    \begin{tabular}{cccc}  \toprule
          & text encoder & image encoder & mIoU (\%) \\\midrule
    \multirow{3}[0]{*}{lr multi} & \cellcolor{Gray}0.0     & \cellcolor{Gray}0.1   & \cellcolor{Gray}\textbf{43.5} \\
          & 0.0     & 1.0     & 42.6 \\
          & 0.1   & 0.1   & 42.2 \\\bottomrule
    \end{tabular}%
  \label{tab:lr}%
\end{table}%

\paragrapha{Effects of optimization of the textual contexts.} Previous works~\cite{coop,clip-adapter} on transferring CLIP models to downstream classification tasks have clearly shown the importance of adapting the textual contexts for different datasets and tasks. We show the effects of optimizing the textual contexts compared to the original prompting strategy proposed in~\cite{clip} in Table~\ref{tab:coop}. We see that although the learnable contexts will introduce additional computation during training (gradient computation for the text encoder), this strategy can bring notable improvement over the baseline. Therefore, we choose to add the learnable textual contexts for our models.

\begin{table}[h]
    \centering
    \caption{Effects of optimization of the textual contexts. We compare the results of using the context optimization~\cite{coop} with directly constructing textual prompts from human defined template and find learnable textual contexts can bring notable improvements. The configuration used in our final models is highlighted in gray.} \small
    \begin{tabular}{cc} \toprule
       Textual Context  &  mIoU (\%)\\\midrule
    \texttt{a photo of a [CLS].}     & 42.9 \\
    \rowcolor{Gray}  CoOp~\cite{coop} & \textbf{43.5} \\\bottomrule
    \end{tabular}
    \label{tab:coop}
\end{table}

\paragrapha{Effects of $\bm{\gamma}$.} Table~\ref{tab:gamma} shows the effects of $\bm{\gamma}$. We see a learnable  $\bm{\gamma}$ initialized with small values can improve the final performance.

\begin{table}[h]
  \centering
  \caption{Ablation study of the residual coefficient $\bm{\gamma}$. The configuration used in our final models is highlighted in gray.} \small
    \begin{tabular}{ccc} \toprule
    initial value of $\bm{\gamma}$ & $\bm{\gamma}$ learnable & mIoU (\%) \\\midrule
    $10^{-4}$ & \xmark     & 42.6 \\
    \rowcolor{Gray} $10^{-4}$ & \cmark     & \textbf{43.5} \\
    1.0 & \cmark     & 42.8 \\\bottomrule
    \end{tabular}%
  \label{tab:gamma}%
\end{table}%

\end{document}